\documentclass[10pt,twocolumn,letterpaper]{article}

\usepackage{iccv}
\usepackage{times}
\usepackage{epsfig}
\usepackage{graphicx}
\usepackage{amsmath}
\usepackage{amssymb}

\usepackage{multirow}
\usepackage{booktabs}
\usepackage{color, colortbl}
\usepackage{pifont}
\usepackage{booktabs}
\usepackage{float}
\usepackage{dsfont}
\usepackage{makecell}

\usepackage{color, colortbl}
\definecolor{Gray}{gray}{0.9}
\definecolor{demphcolor}{RGB}{144,144,144}
\definecolor{airforceblue}{rgb}{0.36, 0.54, 0.66}
\newcommand{\demph}[1]{\textcolor{demphcolor}{#1}}

\makeatletter\renewcommand\paragraph{\@startsection{paragraph}{4}{\z@}
  {.5em \@plus1ex \@minus.2ex}{-.5em}{\normalfont\normalsize\bfseries}}\makeatother
\usepackage[pagebackref=true,breaklinks=true,letterpaper=true,colorlinks,bookmarks=false]{hyperref}

\iccvfinalcopy % *** Uncomment this line for the final submission

\def\iccvPaperID{2853} % *** Enter the ICCV Paper ID here

\ificcvfinal\fi

\begin{document}

%%%%%%%%% TITLE
\title{Transferable Decoding with Visual Entities for Zero-Shot Image Captioning}

% \author{First Author\\
% Institution1\\
% Institution1 address\\
% {\tt\small firstauthor@i1.org}
% % For a paper whose authors are all at the same institution,
% % omit the following lines up until the closing ``}''.
% % Additional authors and addresses can be added with ``\and'',
% % just like the second author.
% % To save space, use either the email address or home page, not both
% \and
% Second Author\\
% Institution2\\
% First line of institution2 address\\
% {\tt\small secondauthor@i2.org}
% }
\author{Junjie Fei$^{1*}$,\ \ Teng Wang$^{1, 2*}$,\ \ Jinrui Zhang$^{1}$,\ \ Zhenyu He$^{3}$,\ \ Chengjie Wang$^{4, 5}$,\ \ Feng Zheng$^{1\dagger}$ \\
$^{1}$Southern University of Science and Technology\ \ $^{2}$The University of Hong Kong\\ $^{3}$Harbin Institute of Technology (Shenzhen)\ \ $^{4}$Tencent\ \ $^{5}$Shanghai Jiao Tong University \\
\tt\small junjiefei@outlook.com\ \ tengwang@connect.hku.hk\ \ zhangjr2018@mail.sustech.edu.cn\\
\tt\small zhenyuhe@hit.edu.cn\ \ jasoncjwang@tencent.com\ \ f.zheng@ieee.org}

\maketitle
% Remove page # from the first page of camera-ready.
\ificcvfinal\thispagestyle{empty}\fi

\begin{abstract}

\let\thefootnote\relax\footnotetext{$^*$ Equal contribution $^{\dagger}$ Corresponding author}
Image-to-text generation aims to describe images using natural language. Recently, zero-shot image captioning based on pre-trained vision-language models (VLMs) and large language models (LLMs) has made significant progress.
However, we have observed and empirically demonstrated that these methods are susceptible to modality bias induced by LLMs and tend to generate descriptions containing objects (entities) that do not actually exist in the image but frequently appear during training (\ie, object hallucination).
In this paper, we propose ViECap, a transferable decoding model that leverages entity-aware decoding to generate descriptions in both seen and unseen scenarios. ViECap incorporates entity-aware hard prompts to guide LLMs' attention toward the visual entities present in the image, enabling coherent caption generation across diverse scenes. With entity-aware hard prompts, ViECap is capable of maintaining performance when transferring from in-domain to out-of-domain scenarios.
Extensive experiments demonstrate that ViECap sets a new state-of-the-art cross-domain (transferable) captioning and performs competitively in-domain captioning compared to previous VLMs-based zero-shot methods. Our code is available at:~\url{https://github.com/FeiElysia/ViECap}

\end{abstract}
\section{Introduction}

\begin{figure}[t]
\begin{center}
   \includegraphics[width=1.0\linewidth]{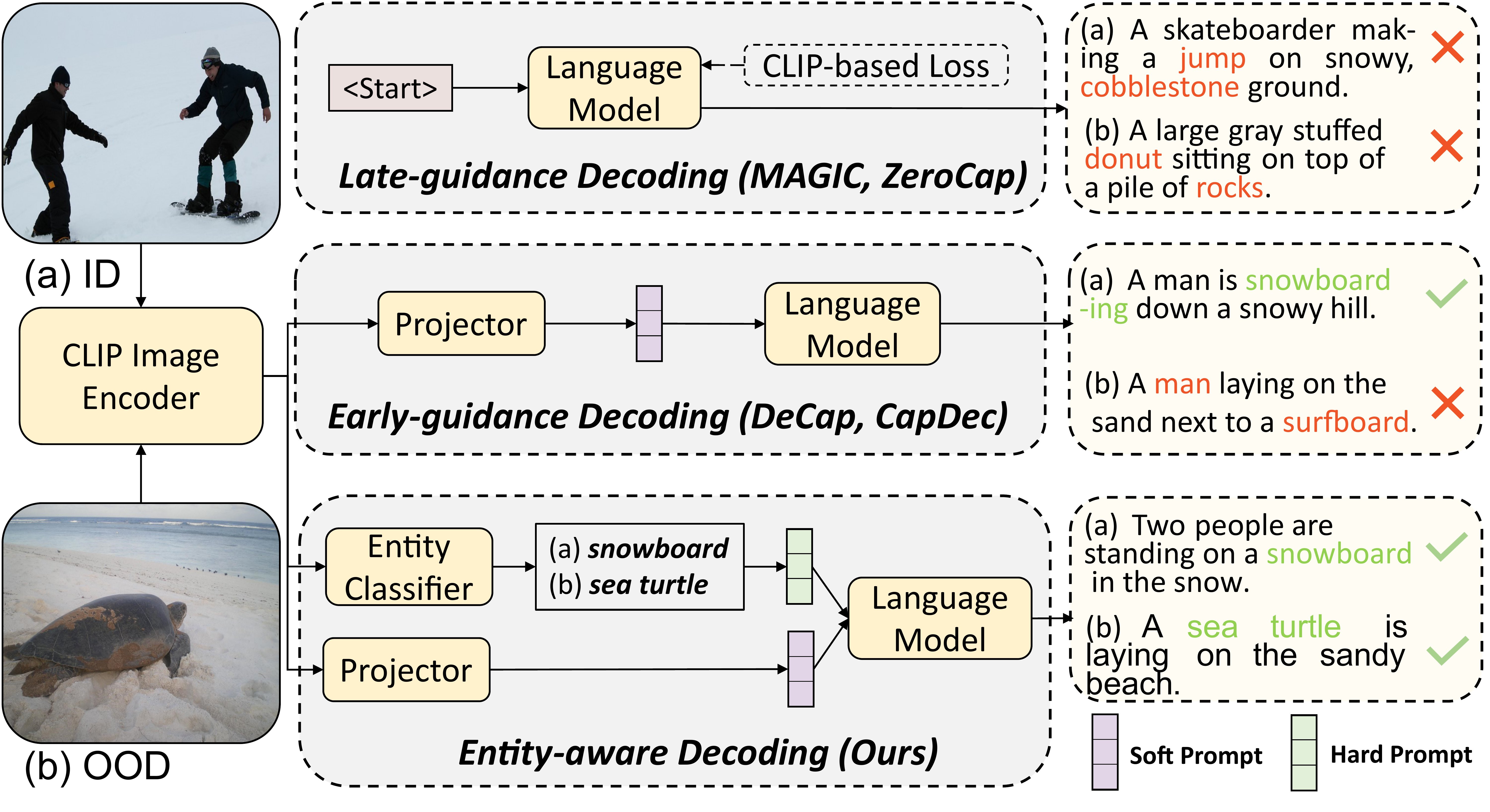}
\end{center}
   \caption{Decoding paradigm for zero-shot image-to-text generation. 
   (a) in-domain (ID) image, (b) out-of-domain (OOD) image. ID refers to objects appearing in the image included in the training corpus, while OOD indicates that they are not included. \color{red}{\ding{53}} \color{black}and \color{green}{\ding{51}} \color{black}refers to the incorrect and correct predictions, respectively.
   Late-guidance methods generate descriptions irrelevant to the image, \eg, \textit{``jump"} and \textit{``donut"}, while early-guidance models often tend to hallucinate objects that are not actually present in the OOD image, \eg, \textit{``surfboard"}. In contrast, our model utilizes entities as additional prompts to describe novel objects in the image, leading to superior transferability in OOD settings, \eg, \textit{``sea turtle''}.}
\label{fig:intro}
\end{figure}

Large-scale pre-trained vision-language models (VLMs) like CLIP~\cite{CLIP} and ALIGN~\cite{ALIGN} showcase impressive zero-shot transferability in various discriminative downstream tasks (\eg, classification~\cite{CLIP}, segmentation~\cite{Lseg,GroupViT}, and detection~\cite{ViLD,GLIP}). However, effectively adapting these pre-trained VLMs into zero-shot generative tasks (\eg, text and image generation) remains an open question that requires further exploration.
Recently, some works~\cite{ZeroCap,MAGIC} have leveraged large language models (LLMs), \eg, GPT~\cite{GPT2,GPT3}, to achieve CLIP-based zero-shot image-to-text generation. They follow a late-guidance paradigm where visual information is injected after completing word prediction. However, the weak visual guidance in this paradigm often results in modality bias, \ie, the language prior in LLMs dominates the decoding process and therefore generates descriptions that are unrelated to the corresponding images.
Fig.~\ref{fig:intro}(a) shows incorrect predictions made by late-guidance decoding, \eg, even if \textit{``jump''} and \textit{``cobblestone''} are unrelated to the image, they finally appear in the predictions due to their close association with predicted words \textit{``skateboarder''} and \textit{``snowy"}.
Similarly, another example in Fig.~\ref{fig:intro}(b) shows that \textit{``donut''} and \textit{``rocks''} are primarily generated by language prior instead of visual guidance. 

Early-guidance methods~\cite{DECAP,CapDec,SMs} provide explicit guidance for word generation in LLMs by prefixing visual prompts to the text tokens. Typically, visual prompts are projected from the CLIP image embedding using a learnable projector.
This early-guidance paradigm significantly alleviates the modality bias and boosts the alignments between the image and the generated captions. 
However, the learnable (soft) visual prompts are prone to overfitting when trained on a limited corpus, leading to poor performance in describing a diverse range of objects (visual entities). This, in turn, may cause object hallucination in the generated captions.
Specifically, when transferring these models to unseen scenarios beyond the training corpus, novel entities are often misrecognized as similar entities frequently appearing in the training corpus. 
As Fig.~\ref{fig:intro} shows, early-guidance decoding is capable of understanding in-domain (ID) images but tends to hallucinate entities that do not actually exist in out-of-domain (OOD) images (\ie, hallucinating \textit{``sea turtle''} with \textit{``surfboard''}, where \textit{``surfboard''} frequently appears in the training corpus). Consequently, the transferability of the well-learned CLIP latent space is degraded into current decoding strategies, significantly limiting their applicability in real-world scenarios. 
We further validate the observed modality bias and object hallucination issues when adapting pre-trained VLMs and LLMs for image-to-text generation through experiments in Sec.~\ref{sec:sec3}.

To address the observed issues, we propose ViECap, which incorporates entity-aware hard prompts to compensate for the degradation of the CLIP latent space caused by learning soft prompts on a specific training corpus. This method is motivated by our observation that the CLIP-based entity classifier can accurately classify both ID and OOD images (\eg, \textit{``snowboard''} and \textit{``sea turtle''} in Fig.~\ref{fig:intro}).
The entity-aware hard prompts enable transferable language decoding from the CLIP latent space. Fig.~\ref{fig:intro} shows that the proposed entity-aware decoding approach is capable of describing both seen and unseen entities in diverse images.
Specifically, ViECap builds on early-guidance decoding methods, \eg, CapDec~\cite{CapDec}.
% Specifically, ViECap builds on CapDec~\cite{CapDec}, an early-guidance decoding method.
Unlike these models, which can only describe entities present in the training corpus, our model can generate captions in diverse scenarios.
Following CapDec, we train ViECap only using text data. The entity-aware hard prompt is the critical design enabling the transferability of our model to diverse captioning scenarios. The hard prompts, constructed by nouns extracted from texts during training or entities retrieved from images during inference, can prompt the LLMs to attend training-agnostic entities based on open vocabulary retrieval through CLIP.
As we find that a naive integration of entities pushes ViECap to learn a copy-then-paste shortcut (\ie, directly copying the entities to captions), we introduce a simple yet efficient entity masking strategy when incorporating the entity-aware hard prompts into early-guidance decoding.

We extensively evaluate ViECap on four benchmarks, NoCaps~\cite{NoCaps}, COCO~\cite{MSCOCO,MSCOCO1}, Flickr30k~\cite{Flickr30k}, and FlickrStyle10K~\cite{StyleNet}. The experimental results demonstrate that ViECap outperforms all other text-only methods and sets a new state-of-the-art in the cross-domain (transferable) setting while remaining competitive with them in the in-domain setting. 
In out-of-domain scenarios (NoCaps), we achieve a margin of 39.2 and 36.3 improvements, respectively, compared to DeCap and CapDec. We even surpass some supervised methods, indicating our model generalizes well to novel entities.
In the experiment on FlickrStyle10K, ViECap effectively generates captions in different styles corresponding to the styles of the training set.
Additionally, the data-efficient experiment shows ViECap's applicability in low-data settings, further highlighting its versatility and effectiveness across various scenarios.

To summarize, our contributions are as follows: 
% 1) We identify modality bias and object hallucination that appears in CLIP-based zero-shot captioning assisted by LLMs, providing timely insights for the generalizability problem.
1) We shed light on the observations and underlying reasons behind the degraded generalizability when adapting pre-trained VLMs and LLMs into image-to-text generation, \ie, modality bias and object hallucination, providing timely and valuable insights for pre-trained large-scale model adaptation.
2) We introduce entity-aware decoding to improve the transferability of zero-shot captioning. Specifically, aided by VLMs, we integrate entity-aware hard prompts with entity masking strategy into the decoding process, guiding LLMs to attend both seen and unseen entities.
3) Extensive experiments show the remarkable zero-shot transferability of ViECap, even in low-data scenarios.
\section{Related Work}
\begin{figure*}[h]
\begin{center}
   \includegraphics[width=1.0\linewidth]{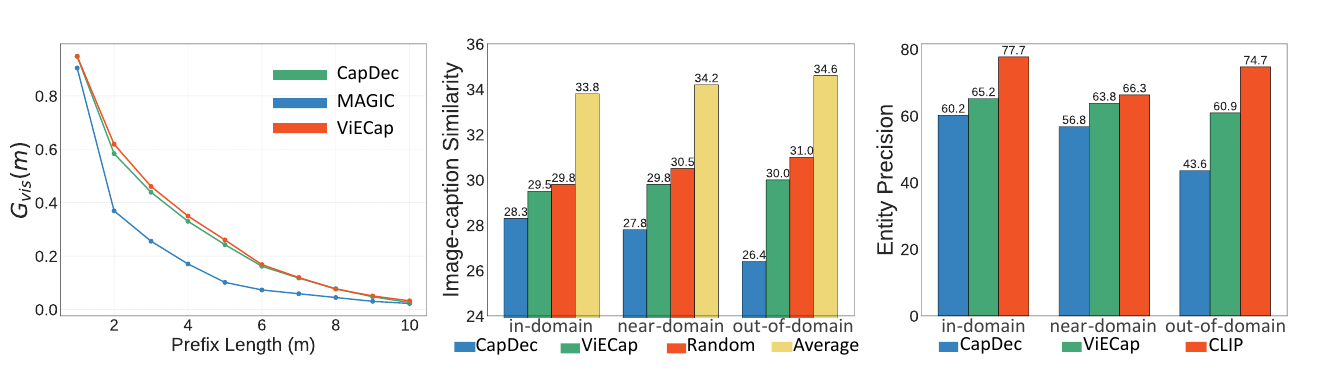}
\end{center}
   \caption{\textbf{Left}: Visual guidance with varied $m$ on COCO. \textbf{Middle}: CLIP similarity between image and captions on NoCaps. ``Random" refers to randomly sampling one ground-truth caption to calculate similarity scores with the paired image, and ``Average" involves calculating the similarity scores by averaging the embeddings of all ground-truth captions. \textbf{Right}: Precision of detected entities using CLIP and captioning models on NoCaps. For both CapDec and ViECap, we evaluate the precision of entity words in the generated captions.
   }
\label{fig:prelim}
\end{figure*}

\paragraph{Supervision in Image Captioning.}
We classify image captioning models into supervised and unsupervised methods based on whether the image-text alignment information is provided during training.
Supervised image captioning methods~\cite{ShowAttendTell, alignmentcaption, AoA, BUTD, MMT, CIDEr, SCA-CNN} are trained with paired (well-aligned) image-text data and typically adopt the encoder-decoder architecture. Initially, diverse vision backbones (\eg, CNN~\cite{ResNet}, ViT~\cite{ViT}) are utilized to extract visual features, which are then fed into a language decoder (\eg, LSTM~\cite{LSTM}, Transformer~\cite{Transformer}) to generate coherent sentences. Various attention mechanisms~\cite{ShowAttendTell, AoA, ICSemanticAttention, SCA-CNN, KnowingWhenToLook} are commonly designed to capture vision-language alignment cues. However, the high cost associated with collecting paired image-text data limits the applicability of these models.
In contrast, Unsupervised image captioning methods~\cite{laina2019towards, feng2019unsupervised} train the model using unpaired image-text data and primarily rely on visual concepts as anchor points to establish pseudo image-text alignment.
Our proposed approach, on the other hand, only requires text data for model training. Compared to previous methods, our method further reduces the data collection cost while exhibiting superior efficiency by eliminating the need for image encoding during training.

\paragraph{Zero-shot Image Captioning.}
Zero-shot image captioning aims to generate image captions without relying on human-annotated data~\cite{DECAP}. Some methods~\cite{CC12M, SimVLM} in this area pre-train the model on large-scale weak image-text pairs and then evaluate the model on target benchmarks without further fine-tuning. 
Another set of methods~\cite{ZeroCap, CapDec, DECAP, MAGIC, SMs} achieves zero-shot image captioning by combining large VLMs and LLMs. Specifically, VLMs provide vision-aware language guidance, which guides LLMs to generate image-related captions. We divide these methods into two paradigms: 1) late-guidance methods (ZeroCap~\cite{ZeroCap} and MAGIC~\cite{MAGIC}) inject visual guidance after word prediction, and 2) early-guidance methods (SMs~\cite{SMs}, CapDec~\cite{CapDec}, and DeCap~\cite{CapDec}) retain visual information in several tokens using VLMs, prompting LLMs to generate image-aware words.
Compared with these methods, we follow the early-guidance paradigm but integrate additional entity-aware hard prompts with an entity masking strategy, which significantly reduces the problem of object hallucination when describing images containing novel objects.

\paragraph{Novel Object Captioning}
This task aims to generate descriptions for images containing unseen objects during training~\cite{MindEye, ConstrainedBeamSearch, NoCaps, NeuralBabyTalk, DCC, NOC}. DCC~\cite{DCC} and NOC~\cite{NOC} leverage object recognition networks to recognize novel concepts. Other methods rely on object detectors (\eg, Faster R-CNN~\cite{FasterRCNN}, Mask R-CNN~\cite{MaskRCNN}) to recognize unseen entities in images~\cite{NeuralBabyTalk, Oscar, VinVL}.
Recently, captioning models leveraging the CLIP-based entity classifier~\cite{SMALLCAP,UniversalCaptioner} have shown even more promising performance in describing novel concepts in images. Despite their success, these methods are trained on limited image-text pairs, making data collection challenging and rendering them susceptible to overfitting to the text style of the training corpus. Consequently, their capability to generate diverse sentences is restricted.
In this study, we extend novel object captioning to a more data-efficient setting. Unlike the aforementioned methods, our model can seamlessly adapt to a new domain by simply fine-tuning with text-only data, thereby improving its transferability and diversity.
\section{Empirical Observations}
\label{sec:sec3}

\begin{figure*}[h]
\begin{center}
   \includegraphics[width=0.95\linewidth]{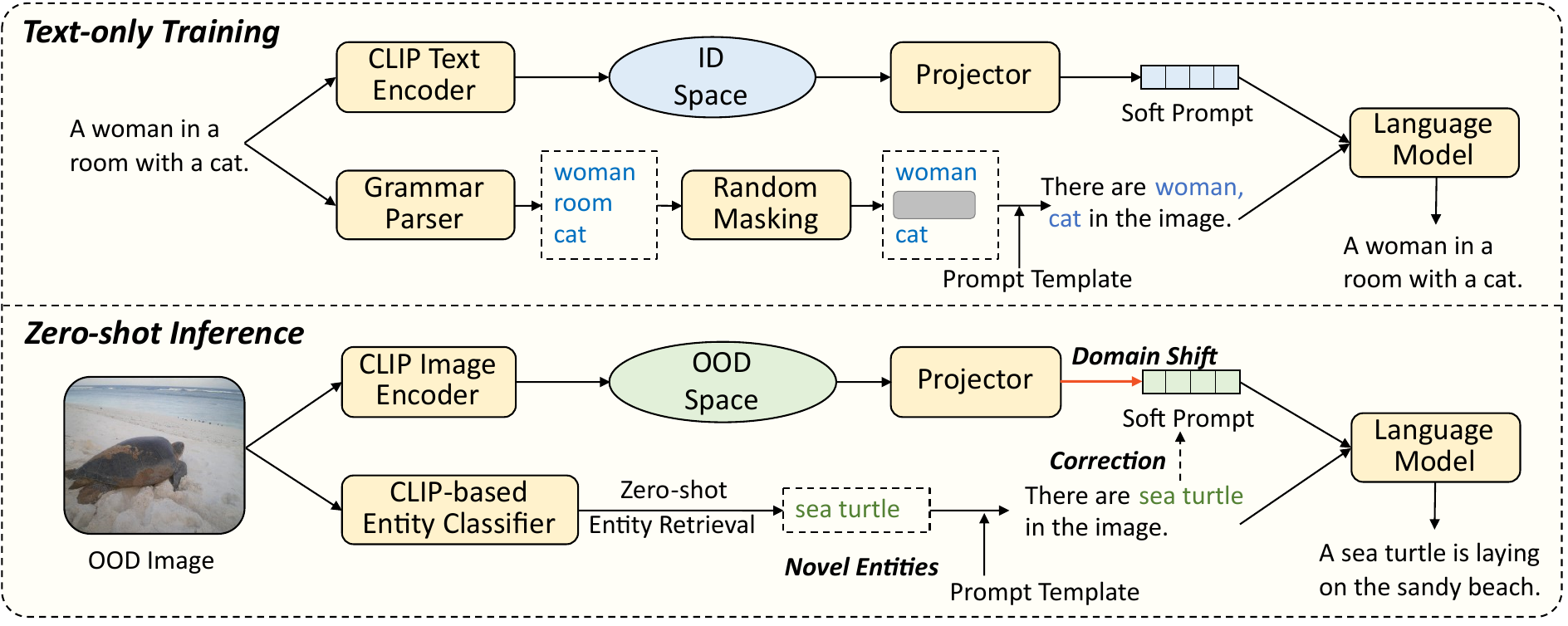}
\end{center}
   \caption{The overview of the proposed ViECap framework. During training, with text-only corpus, nouns are extracted from the sentence by a grammar parser to construct the hard prompt. Then, the soft prompt encodes the overall contexts of the sentence by CLIP text encoder followed by a learnable projector. Two types of prompts are concatenated together as the input for the language model to predict captions. During inference, given a test image, we input the CLIP image embedding into the projector to obtain the soft prompt and introduce a CLIP-based entity classifier to construct the entity-aware hard prompt. With the strong transferability from the training-agnostic hard prompt, our framework is robust to the shift of image domain, achieving excellent captioning performance in both ID and OOD images.}
\label{fig:overview}
\end{figure*}

This section demonstrates the existence of modality bias and object hallucination when adapting VLMs and LLMs for image-to-text generation. It serves as a starting point for the proposed ViECap, which can address such limitations. 

\paragraph{Modality Bias.} 
A good captioning model should strike a balance between visual guidance and language contexts. To evaluate the influence of visual guidance, we design a two-stage decoding strategy: first, we use the captioning model (\eg, MAGIC~\cite{MAGIC}, CapDec~\cite{CapDec}) to generate the first $m$ words of the caption, then we feed these prefix words into a pre-trained language model to obtain the subsequent words based on pure language contexts. The accuracy of generated captions, measured by CIDEr~\cite{CIDEr}, is denoted as CIDEr($m$). We define the importance of visual guidance $G_{\rm vis}(m)$ as: 
\begin{equation}
    G_{\rm vis}(m) = 1 - G_{\rm lang}(m) = 1 - \frac{{\rm CIDEr}(m)}{\rm CIDEr_{model}}
\end{equation}
where $\rm CIDEr_{model}$ is the accuracy of captions generated without a pure language model (\ie, $m$ equals sentence length). $G_{\rm lang}(m)$ is the importance of language contexts. 

If a captioning model is dominated by language priors, $G_{\rm vis}(m)$ will be small as a pure language model can accurately predict captions. As Fig.~\ref{fig:prelim} (left) shows, the late-guidance method MAGIC gains much lower $G_{\rm vis}(m)$ compared to early-guidance methods CapDec and our ViECap, especially when $m$ is small. This observation confirms modality bias towards language in late-guidance methods. 

\paragraph{Object Hallucination.}
While early-guidance decoding alleviates the problem of modality bias effectively, previous models still show limited generalizability towards OOD images containing novel concepts. To illustrate the degradation of transferability in current methods, we calculate the cosine similarity using CLIP between the image and the corresponding generated caption. Fig.~\ref{fig:prelim} (middle) shows that CapDec experiences a gradual performance drop when transferring from ID to OOD settings, while our ViECap exhibits a more robust capability in describing images with different domains.

Object hallucination leads to incorrect entities in the generated caption. We further analyze the precision of entities detected by different methods in Fig.~\ref{fig:prelim} (right). While the CLIP embedding shows remarkable transferability, it is degraded in the caption generation process of CapDec. The accuracy of CapDec drops significantly (60.2 $\rightarrow$ 43.6) when transferring from ID to OOD images.
By explicitly introducing visual entities, ViECap demonstrates the capability to describe both seen and unseen entities in images. Specifically, the accuracy of correctly detecting entities decreases slightly by 4.3 compared to the accuracy predicted by CLIP, which is a reduction of 3.

\section{ViECap}

The proposed ViECap is a transferable captioning framework based on CLIP and trained on a text-only corpus. Specifically, We train a language decoder to decode the CLIP text embedding of sentences and incorporate entity-aware prompts to enable transferable captioning. For zero-shot inference, we directly feed the CLIP image embedding of a given image into the trained decoder to generate captions. Fig.~\ref{fig:overview} illustrates the overall framework of ViECap.

\subsection{Entity-aware Transferable Decoding}
Given the text-only data, our goal is to train an entity-aware language decoder with promising transferability. To this end, we extract two types of visual-aware guidance from the ground-truth caption: 1) nouns in the caption, which serve as anchors for grounding entities in the image. These nouns (\ie, discrete category names) are capable of capturing salient and static visual cues, such as humans, animals, and objects.
2) CLIP text embedding of the caption, which is implicitly aligned with the image embedding, provides the overall contexts across all images, such as scenes and interactions between objects.
We transform entities and the text embedding into prompt tokens to guide the language model (\ie, GPT-2) in predicting captions. During training, we freeze the parameters of the CLIP text encoder to maximize its transferability. We train the projector from scratch and fine-tune the language model using an auto-regressive loss (details can be found in the Appendix).

\begin{table*}
\begin{center}
\small
\setlength{\tabcolsep}{1.5mm}{
\begin{tabular}{l|c|cc|cc|cc|cc}
\toprule
\multicolumn{10}{c}{\textbf{COCO $\Rightarrow$ NoCaps val}} \\
\midrule
\multirow{2}{*}{Methods} & \multirow{2}{*}{Pre-trained Model} & \multicolumn{2}{c|}{in-domain} & \multicolumn{2}{c|}{near-domain} & \multicolumn{2}{c|}{out-of-domain} & \multicolumn{2}{c}{Overall} \\
~  & ~ & CIDEr & SPICE & CIDEr & SPICE & CIDEr & SPICE & CIDEr & SPICE \\
\midrule
\multicolumn{10}{l}{\demph{\textbf{Paired image-text training, zero-shot inference}}} \\
\demph{OSCAR$_{\rm Base}$}~\cite{Oscar} \demph{$_\text{\rm ECCV'20}$} & \demph{\ \ \ Faster R-CNN + BERT$_{\rm Base}$\ \ \ } & \demph{79.6} & \demph{12.3} & \demph{66.1} & \demph{11.5} & \demph{45.3} & \demph{9.7} & \demph{63.8} & \demph{11.2} \\
\demph{ClipCap}~\cite{ClipCap} \demph{$_\text{\rm ArXiv'21}$} & \demph{ViT-B/32 + GPT-2$_{\rm Large}$} & \demph{84.9} & \demph{12.1} & \demph{66.8} & \demph{10.9} & \demph{49.1} & \demph{9.6} & \demph{65.8} & \demph{10.9} \\
\demph{I-Tuning$_{\rm Base}$\ \ \ }~\cite{ITuning} \demph{$_\text{\rm ICASSP'23}$} & \demph{ViT-B/16 + GPT-2$_{\rm Base}$} & \demph{83.9} & \demph{12.4} & \demph{70.3} & \demph{11.7} & \demph{48.1} & \demph{9.5} & \demph{67.8} & \demph{11.4} \\
\demph{SmallCap*}~\cite{SMALLCAP} \demph{$_\text{\rm CVPR'23}$} & \demph{ViT-B/32 + GPT-2$_{\rm Base}$ } & \demph{83.3} & \demph{-} & \demph{77.1} & \demph{-} & \demph{65.0} & \demph{-} & \demph{75.8} & \demph{-} \\
% \midrule
\multicolumn{10}{l}{\textbf{Text-only training, zero-shot inference}} \\
DeCap*~\cite{DECAP} \demph{$_\text{\rm ICLR'22}$} & ViT-B/32 + Transformer & \textbf{65.2} & - & 47.8 & - & 25.8 & - & 45.9 & - \\
CapDec\dag~\cite{CapDec} \demph{$_\text{\rm EMNLP'22}$} & ViT-B/32 + GPT-2$_{\rm Base}$ & 60.1 & 10.2 & 50.2 & 9.3 & 28.7 & 6.0 & 45.9 & 8.3 \\
\rowcolor{Gray}
ViECap \demph{$_\text{\rm ICCV'23}$} & ViT-B/32 + GPT-2$_{\rm Base}$ & 61.1 & \textbf{10.4} & \textbf{64.3} & \textbf{9.9} & \textbf{65.0} & \textbf{8.6} & \textbf{66.2} & \textbf{9.5} \\
\bottomrule
\end{tabular}}
\end{center}
\caption{Cross-domain captioning results on the NoCaps validation set. \dag represents our re-implemented results. * refers to the use of a \textbf{memory bank}. Note that SmallCap reports results on the NoCaps test set, while other methods report results on the NoCaps validation set.}
\label{tab:nocaps}
\end{table*}
\begin{table}
\begin{center}
\small
\setlength{\tabcolsep}{1.2mm}{
\begin{tabular}{l|cccc|cccc}
\toprule
\multirow{2}{*}{Method} &  \multicolumn{4}{c|}{\textbf{COCO $\Rightarrow$ Flickr30k}} & \multicolumn{4}{c}{\textbf{Flickr30k $\Rightarrow$ COCO}}\\
~ & B@4 & M & C & S & B@4 & M & C & S  \\
\midrule
MAGIC~\cite{MAGIC} & 6.2 & 12.2 & 17.5 & 5.9 & 5.2 & 12.5 & 18.3 & 5.7 \\
DeCap~\cite{DECAP} & 16.3 & 17.9 & 35.7 & 11.1 & 12.1 & 18.0 & 44.4 & 10.9 \\
CapDec~\cite{CapDec} & 17.3 & \textbf{18.6} & 35.7 & - & 9.2 & 16.3 & 27.3 & - \\
\rowcolor{Gray}
ViECap & \textbf{17.4} & 18.0 & \textbf{38.4} & \textbf{11.2} & \textbf{12.6} & \textbf{19.3} & \textbf{54.2} & \textbf{12.5} \\
\bottomrule
\end{tabular}}
\end{center}
\caption{Cross-domain captioning results on the Flickr30k test set and COCO test set. All methods in this table use text-only training.}
\label{tab:crossdomain}
\end{table}

\paragraph{Hard Prompt.} 
We first construct a vocabulary of entities, denoted as $\mathcal{V}$. Nouns in the caption, regarded as visual entities, are recognized by the NLTK grammar parser and filtered by this vocabulary. The extracted entities are then inserted into a prompt template ``There are {$e_1$}, ..., {$e_N$} in the image.'', where $e_n$ refers to the $nth$ entity. The entity-aware hard prompt is constructed by a training-agnostic module, enabling strong robustness to the domain shift from ID to OOD images.

\paragraph{Soft Prompt.} 
We first inject Gaussian noise into the CLIP text embedding to alleviate the modality gap as indicated in CapDec~\cite{CapDec}. A trainable projector then transforms the CLIP text embedding to generate the soft prompt. The projector is implemented as a lightweight transformer with $L$ learnable queries as in ClipCap~\cite{ClipCap}. The output features of $L$ query tokens are considered as the soft prompt.
% by taking $F_{T}'$ as input during the training phase. During inference, the adapter encodes $F_{I}$ to generate soft prompts, which can more effectively focus on modeling attributes and relationships as the hard prompts provide explicit visual entities. By doing so, object hallucinations can be reduced significantly in the unseen scene.

\paragraph{Entity masking.} 
We observe that naively integrating nouns to construct hard prompts tends to learn a copy-then-paste shortcut during training, where all nouns are input together to generate a caption, \ie, the model simply pastes the input nouns without making any modification. Consequently, the captioning prediction task becomes trivial, and the model's generalizability is severely impaired, particularly when confronting incorrect entities during inference.
To address this issue, we propose a simple yet effective entity masking strategy that randomly drops a certain proportion of nouns with the masking ratio $r_{mask}$ during training. This strategy significantly alleviates the learning collapse and boosts captioning performance in both ID and OOD settings. The effectiveness of the masking strategy is verified in Tab.~\ref{tab:masking}.

\subsection{Zero-shot Inference}

Once the decoder is trained, we can leverage it for zero-shot captioning inference. Given a test image, we first extract its visual embedding using the CLIP image encoder. We then employ the trained projector to convert the visual embedding into the soft prompt.
For the hard prompt, we again use visual embedding for entity classification. Specifically, we use the manual template ``A photo of \{entity\}" as the entity description for each category in $\mathcal{V}$. Then we rank and select the top $M$ entities with the highest similarity scores between different entity descriptions and the visual embedding to construct the entity-aware hard prompt. Finally, the soft prompt and hard prompt are concatenated together in sequential order and input into the language model to predict the caption auto-regressively.

It should be noted that there exists a training-inference gap in the model structure. Two strategies during training are adopted to address this gap and improve the model performance. Firstly, we use noisy text embedding to bridge the gap between visual and text embedding. Secondly, we propose a non-trivial entity masking mechanism to avoid the copy-then-paste shortcut, meanwhile pushing the model to recover the missing entities from soft prompts.

\paragraph{Transferability on OOD images.}
Trained on limited ID data, the projector may overfit to the ID dataset, leading to a significant performance degradation of the soft prompt for OOD inputs. In contrast, the entity-aware hard prompt, predicted by the frozen CLIP, inherits the powerful transferability from CLIP embeddings. The GPT could flexibly combine the soft and entity-aware hard prompts for a better trade-off between ID and OOD performance.

\section{Experiments}
We conduct extensive experiments to evaluate the performance of ViECap in diverse zero-shot image captioning settings, including 1) cross-domain captioning, 2) in-domain captioning, and 3) data-efficient captioning. The experiments are organized as follows: In Sec.~\ref{sec:sec51}, we assess the transferability of ViECap through the cross-domain setting. Here, the model is trained on a corpus from the source domain and evaluated on a target domain. In Sec.~\ref{sec:sec52}, we focus on the generalizability of ViECap under the in-domain scenario, where the model is trained and evaluated on the same dataset. We conduct data-efficient experiments to assess the applicability of our model in low-data scenarios in Sec.~\ref{sec:sec53}. In Sec.~\ref{sec:sec54}, we perform various ablation experiments to assess the effectiveness of entity-aware decoding. Furthermore, we qualitatively evaluate ViECap in Sec.~\ref{sec:sec55}.

\paragraph{Implementation Details.} We use CLIP-ViT-B/32 as our backbone. The language model is GPT-2$_{\rm base}$ implemented by Wolf et al.~\cite{hugface}. The projector comprises an 8-layer transformer with 8 attention heads and a hidden size of 768. The length of learnable soft prompts is set to 10. During training, we freeze the CLIP text encoder and only train GPT-2 and projector using AdamW~\cite{AdamW} optimizer for all experiments. For caption generation, we use beam search with a beam size of 5. Details are shown in the Appendix.

\paragraph{Datasets and Metrics.} We conduct experiments on four widely used image captioning benchmarks, \ie, NoCaps~\cite{NoCaps}, COCO~\cite{MSCOCO,MSCOCO1}, Flikcr30k~\cite{Flickr30k}, and FlickrStyle10K~\cite{StyleNet}. For COCO and Flickr30k, we follow the commonly used Karpathy split~\cite{alignmentcaption}. For NoCaps, we train our model on the COCO training set and report the results on the validation set, as suggested by OSCAR~\cite{Oscar}. As for FlickrStyle10K,  we follow MemCap~\cite{MemCap}, randomly sampling 6,000 captions as our training set and using the remaining image-text pairs for testing. We report results with common used captioning metrics BLEU@n (B@n)~\cite{BLEU}, METEOR (M)~\cite{METEOR}, CIDEr (C)~\cite{CIDEr} and SPICE (S)~\cite{SPICE}. Refer to the Appendix for details about these datasets.

\paragraph{Methods.} We include several captioning methods as follows: 1) BUTD~\cite{BUTD} and OSCAR~\cite{Oscar} as classic supervised methods, 2) ClipCap~\cite{ClipCap}, I-Tuning~\cite{ITuning}, and SmallCap~\cite{SMALLCAP} as lightweight paired captioning methods that utilize GPT-2 for CLIP-based captioning, 3) ZeroCap~\cite{ZeroCap} as a training-free method, 4) StyleNet~\cite{StyleNet} and MemCap~\cite{MemCap} as classic methods for style captioning, and 5) MAGIC \cite{MAGIC}, CapDec \cite{CapDec}, and DeCap \cite{DECAP} as text-only training methods, which are in line with our work.
Specifically, MAGIC employs late-guidance decoding. CapDec and DeCap adopt early-guidance decoding, which learns soft prompts for caption generation. Notably, DeCap leverages an additional memory bank, and CapDec serves as our baseline.

\begin{table}
\begin{center}
\small
\setlength{\tabcolsep}{0.5mm}{
\begin{tabular}{l|cccc|cccc}
\toprule
\multicolumn{9}{c}{\textbf{In-Domain Captioning}} \\
\midrule
\multirow{2}{*}{Method} & \multicolumn{4}{c|}{COCO} & \multicolumn{4}{c}{Flickr30k} \\
~  & B@4 & M & C & S & B@4 & M & C & S \\
\midrule
\multicolumn{9}{l}{\textbf{\demph{Paired image-text training}}} \\
\demph{BUTD}~\cite{BUTD} \demph{$_\text{\rm CVPR'18}$} & \demph{36.2} & \demph{27.0} & \demph{113.5} & \demph{20.3} & \demph{27.3} & \demph{21.7} & \demph{56.6} & \demph{16.0} \\
\demph{OSCAR}~\cite{Oscar} \demph{$_\text{\rm ECCV'20}$} & \demph{36.5} & \demph{30.3} & \demph{123.7} & \demph{23.1} & \demph{-} & \demph{-} & \demph{-} & \demph{-} \\
\demph{ClipCap}~\cite{ClipCap} \demph{$_\text{\rm ArXiv'21}$} & \demph{33.5} & \demph{27.5} & \demph{113.1} & \demph{21.1} & \demph{-} & \demph{-} & \demph{- }& \demph{-} \\
\demph{I-Tuning}~\cite{ITuning} \demph{$_\text{\rm ICASSP'23}$} & \demph{34.8} & \demph{28.3} & \demph{116.7} & \demph{21.8} & \demph{25.2} & \demph{22.8} & \demph{61.5} & \demph{16.9} \\
\demph{SmallCap*}~\cite{SMALLCAP} \demph{$_\text{\rm CVPR'23}$} & \demph{37.0} & \demph{27.9} & \demph{119.7} & \demph{21.3} & \demph{-} & \demph{-} & \demph{-} & \demph{-} \\
\midrule
\multicolumn{9}{l}{\textbf{Text-only training, zero-shot inference}} \\
ZeroCap~\cite{ZeroCap} \demph{$_\text{\rm CVPR'22}$} & 7.0 & 15.4 & 34.5 & 9.2 & 5.4 & 11.8 & 16.8 & 6.2 \\
MAGIC~\cite{MAGIC} \demph{$_\text{\rm ArXiv'22}$} & 12.9 & 17.4 & 49.3 & 11.3 & 6.4 & 13.1 & 20.4 & 7.1 \\
DeCap*~\cite{DECAP} \demph{$_\text{\rm ICLR'22}$} & 24.7 & 25.0 & 91.2 & \textbf{18.7} & 21.2 & \textbf{21.8} & \textbf{56.7} & \textbf{15.2} \\
CapDec~\cite{CapDec} \demph{$_\text{\rm EMNLP'22}$} & 26.4 & \textbf{25.1} & 91.8 & - & 17.7 & 20.0 & 39.1 & - \\
\rowcolor{Gray}
ViECap \demph{$_\text{\rm ICCV'23}$} & \textbf{27.2} & 24.8 & \textbf{92.9} & 18.2 & \textbf{21.4} & 20.1 & 47.9 & 13.6 \\
\bottomrule
\end{tabular}}
\end{center}
\caption{In-domain captioning results on the COCO test set and Flickr30k test set. * denotes using a \textbf{memory bank}. It should be noted that the result of ZeroCap is copied from MAGIC, and the results of OSCAR and I-Tuning are from their base backbone.}
\label{tab:indomain}
\end{table}
\begin{table}
\begin{center}
\small
\setlength{\tabcolsep}{1.0mm}{
\begin{tabular}{l|cccc|cccc}
\toprule
\multirow{2}{*}{Method} & \multicolumn{4}{c|}{\textbf{Romantic}} & \multicolumn{4}{c}{\textbf{Humorous}}\\
~ & B@1 & B@3 & M & C & B@1 & B@3 & M & C \\
\midrule
StyleNet~\cite{StyleNet} & 13.1 & 1.5 & 4.5 & 7.2 & 13.4 & 0.9 & 4.3 & 11.3 \\
MemCap~\cite{MemCap} & 21.2 & 4.8 & 8.4 & 22.4 & 19.9 & 4.3 & 7.4 & 19.4 \\
CapDec~\cite{CapDec} & 21.4 & 5.0 & 9.6 & 26.9 & \textbf{24.9} & 6.0 & 10.2 & 34.1 \\
\rowcolor{Gray}
ViECap & \textbf{25.7} & \textbf{6.5} & \textbf{10.4} & \textbf{33.6} & 24.3 & \textbf{6.5} & \textbf{10.4} & \textbf{35.0} \\
\bottomrule
\end{tabular}}
\end{center}
\caption{In-domain captioning results on the FlickrStyle10K.}
\label{tab:flickrstyle10k}
\end{table}

\subsection{Cross-Domain Captioning}
\label{sec:sec51}

In this section, we demonstrate the transferability of ViECap in cross-domain captioning. We evaluate ViECap's ability to describe novel entities in images by training it on the COCO training set and testing on the NoCaps validation set without any additional fine-tuning.
As Tab.~\ref{tab:nocaps} shows, ViECap outperforms all other text-only methods by a large margin and even achieves competitive performance compared to some supervised methods in the \textit{out-of-domain} and \textit{Overall} setting, indicating that incorporating the entities-aware hard prompt is beneficial for the model to describe unseen entities.
While other methods experience a notable drop in CIDEr score from the \textit{in-domain} to \textit{out-of-domain} setting in NoCaps, ViECap maintains a minimal fluctuation across different domains, showcasing the remarkable transferability of our model.
In real-world scenarios, the target domain of images is typically agnostic, making the evaluation based on the \textit{Overall} results a better reflection of the models' effectiveness in practical applications. Surprisingly, with text-only corpus, ViECap achieves comparable results to supervised methods, obtaining a CIDEr score of 66.2 compared to 63.8 for OSCAR and 65.8 for ClipCap on the \textit{Overall}. Furthermore, ViECap significantly outperforms the unpaired methods DeCap and CapDec by a large margin of 20.3 CIDEr, demonstrating our model can generate captions with stable quality in diverse domains.

Tab.~\ref{tab:crossdomain} showcases results in more cross-domain settings, where ViECap sets a new state-of-the-art performance on all metrics from Flickr30k to COCO and on most metrics from COCO to Flickr30k.

Both Tab.~\ref{tab:nocaps} and Tab.~\ref{tab:crossdomain} demonstrate the remarkable zero-shot transferability of our model. ViECap is capable of describing images that deviate from the distribution of the training sets, as well as those that do not, making it highly useful when applied in real-world scenarios.

% \begin{table*}
% \begin{center}
% \small
% \setlength{\tabcolsep}{3.0mm}{
% \begin{tabular}{c|c|cccc|cccc}
% \hline \hline
% % \multicolumn{13}{c}{$(A)$ \textbf{In-Domain Captioning}} \\
% % \hline
% \multirow{2}{*}{Methods} & \multirow{2}{*}{Datasets} & \multicolumn{4}{c}{MSCOCO|} & \multicolumn{4}{c}{NoCaps} \\
% ~ & ~ & B@4 & M & C & S & In & Near & Out & Overall\\
% \hline
% % \multicolumn{13}{c}{Supervised Method} \\
% % \hline
% Changpinyo et al. & CC3M & - & - & - & - & 29.2 & 27.5 & 37.3 & 29.7 \\
% Changpinyo et al. & CC12M & - & - & - & - & 20.7 & 24.1 & 41.6 & 27.1 \\
% \hline
% ZeroCap & training-free & 2.6 & 11.5 & 14.6 & 5.5 & - & - & - & - \\
% CLIPRe & CC3M & 4.6 & 13.3 & 25.6 & 9.2 & 23.3 & 26.8 & 36.5 & 28.2 \\
% DeCap & CC3M(1M extra memory) & 8.8 & 16.0 & 42.1 & 10.9 & \textbf{34.8} & 37.7 & 49.9 & 39.7 \\
% CapDec\textcolor{red}{(run)} & w/o hard prompt & - & - & - & - & - & - & - & - \\
% \hline
% Ours & CC3M & \textbf{13.2} & \textbf{18.5} & \textbf{51.8} & \textbf{13.0} & 32.4 & \textbf{40.6} & \textbf{50.5} & - \\
% % \multicolumn{13}{c}{Text-only training, zero-shot on image-text pairs} \\
% % \hline
% % Magic & 1.0 & 1.0 & 1.0 & 1.0 & 1.0 & 1.0 & 1.0 & 1.0 & 1.0 & 1.0 & 1.0 & 1.0\\
% \hline \hline
% \end{tabular}}
% \end{center}
% \caption{Results.   Ours is better.}
% \end{table*}

\subsection{In-Domain Captioning}
\label{sec:sec52}

To further assess the generalizability of ViECap, we conduct evaluations on COCO, Flickr30k, and FlickrStyle10K in the in-domain setting, where the training and testing data are from the same dataset. As shown in Tab.~\ref{tab:indomain}, our proposed model outperforms CapDec, our baseline method, in most metrics. We attribute this improvement to our entity-aware hard prompt, which explicitly emphasizes infrequent object concepts, thereby mitigating the long-tail problem associated with the existing dataset.
It is worth noting that DeCap utilizes a large memory bank to bridge the modality gap, which may not be practical in real-world scenarios. In contrast, our approach achieves comparable performance with an acceptable memory complexity, highlighting the effectiveness of the proposed ViECap. Tab.~\ref{tab:flickrstyle10k} shows that ViECap achieves state-of-the-art performance on FlickStyle10K. These results demonstrate that ViECap can adapt well to diverse style text data, showcasing its versatility and strong generalizability.

\begin{table}
\small
\begin{center}
\setlength{\tabcolsep}{2.0 mm}{
\begin{tabular}{r|c|c|cccc}
\toprule
\multirow{2}{*}{Data} & \multirow{2}{*}{Method} &\multicolumn{1}{c|}{\textbf{COCO}}&\multicolumn{4}{c}{\textbf{NoCaps val}} \\
 & & Test & In & Near & Out & Overall \\
\midrule
\multirow{2}{*}{0.1\%} & CapDec & 24.0 & 13.2 & 11.0 & 6.2 & 10.4 \\
% & ViECap & 36.0 & 17.1 & 14.5 & 11.4 & 14.4 \\
& ViECap & \textbf{32.3} & \textbf{20.9} & \textbf{27.6} & \textbf{34.9} & \textbf{30.2} \\
\midrule
\multirow{2}{*}{1\%} & CapDec & 55.8 & 29.6 & 20.5 & 9.8 & 18.9 \\
% & ViECap &64.2& 36.1 & 39.9 & 40.6 & 41.2\\
& ViECap & \textbf{63.9} & \textbf{34.6} & \textbf{39.9} & \textbf{39.3} & \textbf{40.4}\\
\midrule
\multirow{2}{*}{10\%} & CapDec & \textbf{83.6} & 47.3 & 39.8 & 19.1& 35.4\\
& ViECap & 83.4 & \textbf{45.9} & \textbf{51.8} & \textbf{48.7}  & \textbf{53.3} \\
\midrule
% \multirow{2}{*}{20\%} & CapDec & & & & \\
% & ViECap & \\
% \midrule
\multirow{2}{*}{100\%} & CapDec & 92.7 & 60.1 & 50.2 & 28.7 & 45.9  \\
& ViECap & \textbf{92.9} & \textbf{61.1} & \textbf{64.3} & \textbf{65.0} & \textbf{66.2} \\
\bottomrule
\end{tabular}}
\end{center}
\caption{Data-efficient captioning results.}
\label{tab:fewshot}
\end{table}

\subsection{Data-Efficient Captioning}
\label{sec:sec53}

In this section, we explore ViECap's capability to learn from low-data scenarios. Specifically, we randomly sample different scales of data from the COCO training set to train ViECap. For simplicity, we leverage In, Near, and Out to denote \textit{in-domain}, \textit{near-domain}, and \textit{out-of-domain}, respectively.
As shown in Tab.~\ref{tab:fewshot}, ViECap outperforms CapDec across all data scales. Even with as little as 0.1\% of the data, ViECap can still generate reasonable captions and maintain transferability (CIDEr: 32.3 on the COCO testing set vs. 30.2 on the \textit{Overall} of NoCaps validation set), suggesting ViECap is data-efficient and applicable in low-data settings.

\subsection{Ablation Studies}
\label{sec:sec54}

\begin{table}
\small
\begin{center}
\setlength{\tabcolsep}{1.2 mm}{
\begin{tabular}{l|c|cccc}
\toprule
\multirow{2}{*}{Method} & \multicolumn{1}{c|}{\textbf{COCO}}&\multicolumn{4}{c}{\textbf{NoCaps val}} \\
~ & Test & In & Near & Out & Overall \\
\midrule
Baseline & 92.7 & 60.1 & 50.2 & 28.7 & 45.9 \\
\ + Entity & 53.2 & 32.4 & 40.3 & 53.1 & 44.8 \\
\ + Entity + Masking (20\%) & 88.1 & 54.1 & 60.5 & 63.5 & 62.7 \\
\rowcolor{Gray}
\ + Entity + Masking (40\%) & 92.9 & \textbf{61.1} & \textbf{64.3} & \textbf{65.0} & \textbf{66.2} \\
\ + Entity + Masking (60\%) & \textbf{94.6} & 59.1 & 64.0 & 63.9 & 65.5 \\
\ + Entity + Masking (80\%) & 94.5 & 57.8 & \textbf{64.3} & 63.5 & 65.3 \\
\bottomrule
\end{tabular}}
\end{center}
\caption{Ablation studies of entity masking.}
\label{tab:masking}
\end{table}

\begin{table}
\small
\begin{center}
\setlength{\tabcolsep}{1.0 mm}{
\begin{tabular}{l|c|cccc}
\toprule
\multirow{2}{*}{Method} & \multicolumn{1}{c|}{\textbf{COCO}}&\multicolumn{4}{c}{\textbf{NoCaps val}} \\
~ & Test & In & Near & Out & Overall \\
\midrule
Baseline (Soft-only) & 92.7 & 60.1 & 50.2 & 28.7 & 45.9 \\
Entity-only & 61.4 & 29.9 & 41.0 & 49.8 & 44.4 \\
Entity + Soft & 92.5 & 60.8 & 63.6 & \textbf{65.1} & 65.7 \\
Soft + Entity (w/o ensemble) & 92.3 & 57.3 & 59.5 & 59.1 & 61.0 \\
\rowcolor{Gray}
Soft + Entity & \textbf{92.9} & \textbf{61.1} & \textbf{64.3} & 65.0 & \textbf{66.2} \\
\bottomrule
\end{tabular}}
\end{center}
\caption{Ablation studies of prompts.}
\label{tab:prompts}
\end{table}

\begin{figure*}[t]
\begin{center}
   \includegraphics[width=1.0\linewidth]{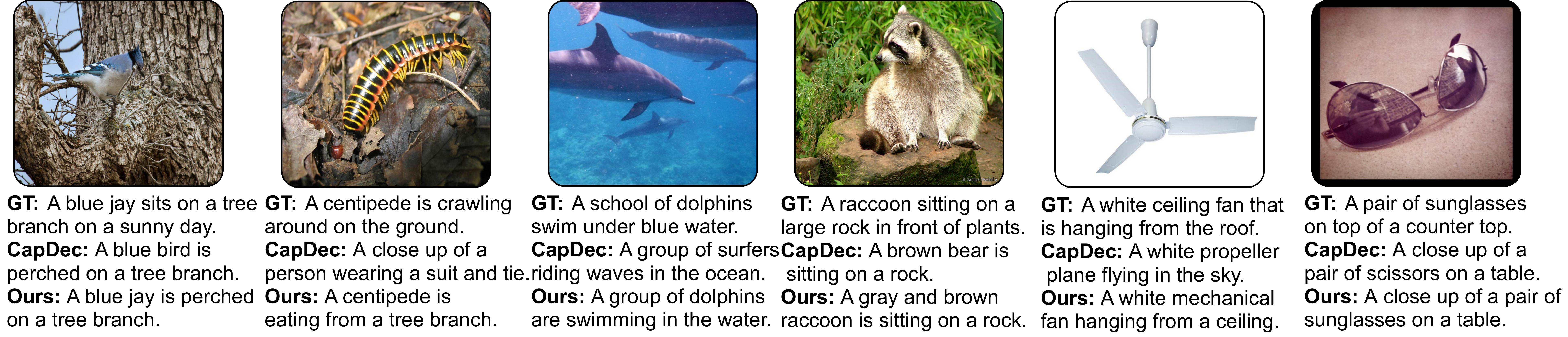}
\end{center}
   \caption{Visualization of generated captions of some images from the out-of-domain setting on the NoCaps validation set, which contain unseen entities. GT refers to the Ground Truth, CapDec and Ours denote caption generated by CapDec and ViECap, respectively.}
\label{fig:visual}
\end{figure*}

We conduct comprehensive ablation studies to explore the influence of entities and prompt structures on our model. Additionally, we evaluate the advantages of using a larger-scale language model for ViECap. In all these experiments, we assess the performance of our model on both the COCO testing set and the NoCaps validation set.

\paragraph{Masking Rate.} We begin by investigating the impact of entities on ViECap by randomly masking entities at different rates. As shown in Tab.~\ref{tab:masking}, incorporating entities enhances our model's ability to perceive unseen entities (from $28.7$ to $53.1$).
However, as mentioned before, the model may learn a shortcut from detected entities, leading to a rapid decline in ID performance (from $92.7$ to $53.2$). The CIDEr score on the COCO testing set gradually increases as the masking rate increases, indicating that entity masking can prevent ViECap from relying heavily on entities.
For the NoCaps validation set, the performance of ViECap first increases and then decreases, showing that a moderate entity masking rate benefits caption prediction in unseen scenarios. The results of this experiment demonstrate that the proposed entity masking strategy boosts the captioning performance of ViECap across diverse scenes.

\paragraph{Prompts.} We then explore the impact of different prompt generation methods on the performance of ViECap. As shown in Tab.~\ref{tab:prompts}, learning only soft prompts leads to overfitting to in-domain captioning, resulting in poor performance when describing novel entities. Incorporating hard prompts improves captioning performance for unseen images, but solely modeling entity information leads to reduced performance on in-domain captioning.
When combining soft and hard prompts, we achieve comparable ID performance with soft prompts-only methods and superior performance on OOD scenarios. Additionally, we find that the order of soft and hard prompts does not affect ViECap's performance.
To construct hard prompts with visual entities, we leverage CLIP-based retrieval, where the accuracy of retrieval benefits from the prompt ensemble~\footnote{The prompt engineering is released by OpenAI.~\url{https://github.com/openai/CLIP/blob/main/notebooks}}.

\begin{table}
\small
\begin{center}
\setlength{\tabcolsep}{0.85 mm}{
\begin{tabular}{l|cc|c|cccc}
\toprule
\multirow{2}{*}{Model} & \multirow{2}{*}{pt.} & \multirow{2}{*}{\#para} & \multicolumn{1}{c|}{\textbf{COCO}}&\multicolumn{4}{c}{\textbf{NoCaps val}} \\
 & & & Test & In & Near & Out & Overall \\
\midrule
\multicolumn{5}{l}{\textbf{Tuned language model}} \\
% GPT$_{\rm Base}$ (4-layer) & $\times$ & 67M & 89.8 & 51.8 & 53.4 & 48.1 & 53.1 \\
GPT$_{\rm Base}$ (4-layer) & $\times$ & 67M & 89.8 & 54.5 & 54.2 & 50.2 & 54.6 \\
GPT$_{\rm Base}$ & $\surd$ &124M & 92.9 & {61.1} & {64.3} & {65.0} & 66.2\\
\midrule
\multicolumn{5}{l}{\textbf{Frozen language model}} \\
GPT$_{\rm Base}$ & $\surd$ &124M & 88.0 & 57.7 & 60.2 & 61.4 & 62.0 \\
GPT$_{\rm Large}$ & $\surd$ &774M & 91.7 & 59.4 & 64.4 & 68.4 & 66.9 \\
GPT$_{\rm XL}$ & $\surd$ &1.5B & 94.5 & 63.4 & 66.6 & 68.9 & 68.9 \\
OPT$_{\rm 1.3B}$ & $\surd$ &1.3B & 95.6 & 64.9 & 68.7 & 69.9 & 70.7 \\
OPT$_{\rm 2.7B}$ & $\surd$ &2.7B & 96.9 & 64.7 & 70.2 & 71.9 & 72.1 \\
% OPT-iml & 1.3B & & & & & \\
\bottomrule
\end{tabular}}
\end{center}
\caption{ViECap with different scales of language models. ``pt." represents using pre-trained weights.}
\label{scaleup}
\end{table}

\paragraph{Scaling up ViECap.} We assess diverse pre-trained LLMs, ranging from GPT-2 to OPT~\cite{OPT}, to investigate the impact of scaling up language models on the capability of ViECap to describe novel entities in images.
The results are shown in Tab.~\ref{scaleup}. As the model parameters increase, the performance of ViECap continues to improve. Notably, we freeze the language model for more effective training. To our surprise, ViECap can effectively leverage the information from the language model without the need for further fine-tuning.

\subsection{Qualitative Evaluation}
\label{sec:sec55}

Fig.~\ref{fig:visual} displays ground truth captions and captions generated by CapDec and ViECap trained on COCO. Images are from the \textit{out-of-domain} set in the NoCaps validation set, which contains unseen entities. In the first image in Fig.~\ref{fig:visual}, ViECap correctly recognizes the specific entity ``jay" while CapDec mistakenly identifies a generic entity ``bird". Similar outcomes can be observed in the other instances shown in the figure.
CapDec exhibits object hallucination in its textual descriptions, while ViECap demonstrates the ability to generate high-quality descriptions of entities in novel scenarios, illustrating the effectiveness of using hard prompts to guide LLMs in attending to entities in images. More visualization results can be found in the Appendix.
\section{Conclusion}

In this paper, we have comprehensively investigated the challenges of adapting pre-trained VLMs and LLMs for image-to-text generation. Our empirical findings demonstrate the existence of modality bias and object hallucination, highlighting the limited transferability when adapting pre-trained models to downstream tasks and providing valuable insight for further research in this area.
We propose an entity-aware decoding approach to address the observed issues. By leveraging the CLIP latent space to prompt GPT-2 during caption generation, our method, ViECap, demonstrates remarkable performance in both seen and unseen scenarios.
Extensive experiments showcase that our ViECap outperforms existing zero-shot methods in transferability and achieves competitive performance on zero-shot in-domain captioning. Moreover, ViECap proves to be data-efficient in low-data settings on COCO. Experiment on FlickrStyle10K shows that our model can also generate captions in different styles based on the training corpus style.

\paragraph{Acknowledgments.} This work was supported by the National Key R\&D Program of China (Grant NO. 2022YFF1202903) and the National Natural Science Foundation of China (Grant NO. 62122035).

{\small
\bibliographystyle{ieee_fullname}
\bibliography{egbib}

\begin{thebibliography}{10}\itemsep=-1pt

\bibitem{NoCaps}
Harsh Agrawal, Karan Desai, Yufei Wang, Xinlei Chen, Rishabh Jain, Mark
  Johnson, Dhruv Batra, Devi Parikh, Stefan Lee, and Peter Anderson.
\newblock Nocaps: Novel object captioning at scale.
\newblock {\em Proceedings of the IEEE/CVF international conference on computer
  vision}, pages 8948--8957, 2019.

\bibitem{ConstrainedBeamSearch}
Peter Anderson, Basura Fernando, Mark Johnson, and Stephen Gould.
\newblock Guided open vocabulary image captioning with constrained beam search.
\newblock {\em arXiv preprint arXiv:1612.00576}, 2016.

\bibitem{SPICE}
Peter Anderson, Basura Fernando, Mark Johnson, and Stephen Gould.
\newblock Spice: Semantic propositional image caption evaluation.
\newblock {\em Computer Vision--ECCV 2016: 14th European Conference, Amsterdam,
  The Netherlands, October 11-14, 2016, Proceedings, Part V 14}, pages
  382--398, 2016.

\bibitem{BUTD}
Peter Anderson, Xiaodong He, Chris Buehler, Damien Teney, Mark Johnson, Stephen
  Gould, and Lei Zhang.
\newblock Bottom-up and top-down attention for image captioning and visual
  question answering.
\newblock {\em Proceedings of the IEEE conference on computer vision and
  pattern recognition}, pages 6077--6086, 2018.

\bibitem{GPT3}
Tom Brown, Benjamin Mann, Nick Ryder, Melanie Subbiah, Jared~D Kaplan, Prafulla
  Dhariwal, Arvind Neelakantan, Pranav Shyam, Girish Sastry, Amanda Askell,
  et~al.
\newblock Language models are few-shot learners.
\newblock {\em Advances in neural information processing systems},
  33:1877--1901, 2020.

\bibitem{CC12M}
Soravit Changpinyo, Piyush Sharma, Nan Ding, and Radu Soricut.
\newblock Conceptual 12m: Pushing web-scale image-text pre-training to
  recognize long-tail visual concepts.
\newblock {\em Proceedings of the IEEE/CVF Conference on Computer Vision and
  Pattern Recognition}, pages 3558--3568, 2021.

\bibitem{SCA-CNN}
Long Chen, Hanwang Zhang, Jun Xiao, Liqiang Nie, Jian Shao, Wei Liu, and
  Tat-Seng Chua.
\newblock Sca-cnn: Spatial and channel-wise attention in convolutional networks
  for image captioning.
\newblock {\em Proceedings of the IEEE conference on computer vision and
  pattern recognition}, pages 5659--5667, 2017.

\bibitem{MSCOCO1}
Xinlei Chen, Hao Fang, Tsung-Yi Lin, Ramakrishna Vedantam, Saurabh Gupta, Piotr
  Doll{\'a}r, and C~Lawrence Zitnick.
\newblock Microsoft coco captions: Data collection and evaluation server.
\newblock {\em arXiv preprint arXiv:1504.00325}, 2015.

\bibitem{MindEye}
Xinlei Chen and C Lawrence~Zitnick.
\newblock Mind's eye: A recurrent visual representation for image caption
  generation.
\newblock {\em Proceedings of the IEEE conference on computer vision and
  pattern recognition}, pages 2422--2431, 2015.

\bibitem{UniversalCaptioner}
Marcella Cornia, Lorenzo Baraldi, Giuseppe Fiameni, and Rita Cucchiara.
\newblock Universal captioner: Inducing content-style separation in
  vision-and-language model training.
\newblock {\em arXiv preprint arXiv:2111.12727}, 2021.

\bibitem{MMT}
Marcella Cornia, Matteo Stefanini, Lorenzo Baraldi, and Rita Cucchiara.
\newblock Meshed-memory transformer for image captioning.
\newblock {\em Proceedings of the IEEE/CVF conference on computer vision and
  pattern recognition}, pages 10578--10587, 2020.

\bibitem{METEOR}
Michael Denkowski and Alon Lavie.
\newblock Meteor universal: Language specific translation evaluation for any
  target language.
\newblock {\em Proceedings of the ninth workshop on statistical machine
  translation}, pages 376--380, 2014.

\bibitem{ViT}
Alexey Dosovitskiy, Lucas Beyer, Alexander Kolesnikov, Dirk Weissenborn,
  Xiaohua Zhai, Thomas Unterthiner, Mostafa Dehghani, Matthias Minderer, Georg
  Heigold, Sylvain Gelly, et~al.
\newblock An image is worth 16x16 words: Transformers for image recognition at
  scale.
\newblock {\em arXiv preprint arXiv:2010.11929}, 2020.

\bibitem{feng2019unsupervised}
Yang Feng, Lin Ma, Wei Liu, and Jiebo Luo.
\newblock Unsupervised image captioning.
\newblock In {\em Proceedings of the IEEE/CVF Conference on Computer Vision and
  Pattern Recognition}, pages 4125--4134, 2019.

\bibitem{StyleNet}
Chuang Gan, Zhe Gan, Xiaodong He, Jianfeng Gao, and Li Deng.
\newblock Stylenet: Generating attractive visual captions with styles.
\newblock {\em Proceedings of the IEEE conference on computer vision and
  pattern recognition}, pages 3137--3146, 2017.

\bibitem{ViLD}
Xiuye Gu, Tsung-Yi Lin, Weicheng Kuo, and Yin Cui.
\newblock Open-vocabulary object detection via vision and language knowledge
  distillation.
\newblock {\em arXiv preprint arXiv:2104.13921}, 2021.

\bibitem{MaskRCNN}
Kaiming He, Georgia Gkioxari, Piotr Doll{\'a}r, and Ross Girshick.
\newblock Mask r-cnn.
\newblock {\em Proceedings of the IEEE international conference on computer
  vision}, pages 2961--2969, 2017.

\bibitem{ResNet}
Kaiming He, Xiangyu Zhang, Shaoqing Ren, and Jian Sun.
\newblock Deep residual learning for image recognition.
\newblock {\em Proceedings of the IEEE conference on computer vision and
  pattern recognition}, pages 770--778, 2016.

\bibitem{DCC}
Lisa~Anne Hendricks, Subhashini Venugopalan, Marcus Rohrbach, Raymond Mooney,
  Kate Saenko, and Trevor Darrell.
\newblock Deep compositional captioning: Describing novel object categories
  without paired training data.
\newblock {\em Proceedings of the IEEE conference on computer vision and
  pattern recognition}, pages 1--10, 2016.

\bibitem{LSTM}
Sepp Hochreiter and J{\"u}rgen Schmidhuber.
\newblock Long short-term memory.
\newblock {\em Neural computation}, 9(8):1735--1780, 1997.

\bibitem{AoA}
Lun Huang, Wenmin Wang, Jie Chen, and Xiao-Yong Wei.
\newblock Attention on attention for image captioning.
\newblock {\em Proceedings of the IEEE/CVF international conference on computer
  vision}, pages 4634--4643, 2019.

\bibitem{ALIGN}
Chao Jia, Yinfei Yang, Ye Xia, Yi-Ting Chen, Zarana Parekh, Hieu Pham, Quoc Le,
  Yun-Hsuan Sung, Zhen Li, and Tom Duerig.
\newblock Scaling up visual and vision-language representation learning with
  noisy text supervision.
\newblock {\em International Conference on Machine Learning}, pages 4904--4916,
  2021.

\bibitem{alignmentcaption}
Andrej Karpathy and Li Fei-Fei.
\newblock Deep visual-semantic alignments for generating image descriptions.
\newblock {\em Proceedings of the IEEE conference on computer vision and
  pattern recognition}, pages 3128--3137, 2015.

\bibitem{AdamW}
Diederik~P Kingma and Jimmy Ba.
\newblock Adam: A method for stochastic optimization.
\newblock {\em arXiv preprint arXiv:1412.6980}, 2014.

\bibitem{laina2019towards}
Iro Laina, Christian Rupprecht, and Nassir Navab.
\newblock Towards unsupervised image captioning with shared multimodal
  embeddings.
\newblock In {\em Proceedings of the IEEE/CVF International Conference on
  Computer Vision}, pages 7414--7424, 2019.

\bibitem{Lseg}
Boyi Li, Kilian~Q Weinberger, Serge Belongie, Vladlen Koltun, and Ren{\'e}
  Ranftl.
\newblock Language-driven semantic segmentation.
\newblock {\em arXiv preprint arXiv:2201.03546}, 2022.

\bibitem{GLIP}
Liunian~Harold Li, Pengchuan Zhang, Haotian Zhang, Jianwei Yang, Chunyuan Li,
  Yiwu Zhong, Lijuan Wang, Lu Yuan, Lei Zhang, Jenq-Neng Hwang, et~al.
\newblock Grounded language-image pre-training.
\newblock {\em Proceedings of the IEEE/CVF Conference on Computer Vision and
  Pattern Recognition}, pages 10965--10975, 2022.

\bibitem{DECAP}
Wei Li, Linchao Zhu, Longyin Wen, and Yi Yang.
\newblock Decap: Decoding clip latents for zero-shot captioning.
\newblock {\em International Conference on Learning Representations}, 2023.

\bibitem{Oscar}
Xiujun Li, Xi Yin, Chunyuan Li, Pengchuan Zhang, Xiaowei Hu, Lei Zhang, Lijuan
  Wang, Houdong Hu, Li Dong, Furu Wei, et~al.
\newblock Oscar: Object-semantics aligned pre-training for vision-language
  tasks.
\newblock {\em Computer Vision--ECCV 2020: 16th European Conference, Glasgow,
  UK, August 23--28, 2020, Proceedings, Part XXX 16}, pages 121--137, 2020.

\bibitem{MSCOCO}
Tsung-Yi Lin, Michael Maire, Serge Belongie, James Hays, Pietro Perona, Deva
  Ramanan, Piotr Doll{\'a}r, and C~Lawrence Zitnick.
\newblock Microsoft coco: Common objects in context.
\newblock {\em Computer Vision--ECCV 2014: 13th European Conference, Zurich,
  Switzerland, September 6-12, 2014, Proceedings, Part V 13}, pages 740--755,
  2014.

\bibitem{KnowingWhenToLook}
Jiasen Lu, Caiming Xiong, Devi Parikh, and Richard Socher.
\newblock Knowing when to look: Adaptive attention via a visual sentinel for
  image captioning.
\newblock {\em Proceedings of the IEEE conference on computer vision and
  pattern recognition}, pages 375--383, 2017.

\bibitem{NeuralBabyTalk}
Jiasen Lu, Jianwei Yang, Dhruv Batra, and Devi Parikh.
\newblock Neural baby talk.
\newblock {\em Proceedings of the IEEE conference on computer vision and
  pattern recognition}, pages 7219--7228, 2018.

\bibitem{ITuning}
Ziyang Luo, Zhipeng Hu, Yadong Xi, Rongsheng Zhang, and Jing Ma.
\newblock I-tuning: Tuning frozen language models with image for lightweight
  image captioning.
\newblock {\em ICASSP 2023-2023 IEEE International Conference on Acoustics,
  Speech and Signal Processing (ICASSP)}, pages 1--5, 2023.

\bibitem{ClipCap}
Ron Mokady, Amir Hertz, and Amit~H Bermano.
\newblock Clipcap: Clip prefix for image captioning.
\newblock {\em arXiv preprint arXiv:2111.09734}, 2021.

\bibitem{CapDec}
David Nukrai, Ron Mokady, and Amir Globerson.
\newblock Text-only training for image captioning using noise-injected clip.
\newblock {\em arXiv preprint arXiv:2211.00575}, 2022.

\bibitem{BLEU}
Kishore Papineni, Salim Roukos, Todd Ward, and Wei-Jing Zhu.
\newblock Bleu: a method for automatic evaluation of machine translation.
\newblock {\em Proceedings of the 40th annual meeting of the Association for
  Computational Linguistics}, pages 311--318, 2002.

\bibitem{CLIP}
Alec Radford, Jong~Wook Kim, Chris Hallacy, Aditya Ramesh, Gabriel Goh,
  Sandhini Agarwal, Girish Sastry, Amanda Askell, Pamela Mishkin, Jack Clark,
  et~al.
\newblock Learning transferable visual models from natural language
  supervision.
\newblock {\em International conference on machine learning}, pages 8748--8763,
  2021.

\bibitem{GPT2}
Alec Radford, Jeffrey Wu, Rewon Child, David Luan, Dario Amodei, Ilya
  Sutskever, et~al.
\newblock Language models are unsupervised multitask learners.
\newblock {\em OpenAI blog}, 1(8):9, 2019.

\bibitem{OfficeHome}
Mohammad~Mahfujur Rahman, Clinton Fookes, Mahsa Baktashmotlagh, and Sridha
  Sridharan.
\newblock Multi-component image translation for deep domain generalization.
\newblock {\em 2019 IEEE Winter Conference on Applications of Computer Vision
  (WACV)}, pages 579--588, 2019.

\bibitem{SMALLCAP}
Rita Ramos, Bruno Martins, Desmond Elliott, and Yova Kementchedjhieva.
\newblock Smallcap: lightweight image captioning prompted with retrieval
  augmentation.
\newblock {\em Proceedings of the IEEE/CVF Conference on Computer Vision and
  Pattern Recognition}, pages 2840--2849, 2023.

\bibitem{FasterRCNN}
Shaoqing Ren, Kaiming He, Ross Girshick, and Jian Sun.
\newblock Faster r-cnn: Towards real-time object detection with region proposal
  networks.
\newblock {\em Advances in neural information processing systems}, 28, 2015.

\bibitem{MAGIC}
Yixuan Su, Tian Lan, Yahui Liu, Fangyu Liu, Dani Yogatama, Yan Wang, Lingpeng
  Kong, and Nigel Collier.
\newblock Language models can see: plugging visual controls in text generation.
\newblock {\em arXiv preprint arXiv:2205.02655}, 2022.

\bibitem{ZeroCap}
Yoad Tewel, Yoav Shalev, Idan Schwartz, and Lior Wolf.
\newblock Zero-shot image-to-text generation for visual-semantic arithmetic.
\newblock {\em arXiv preprint arXiv:2111.14447}, 2021.

\bibitem{Transformer}
Ashish Vaswani, Noam Shazeer, Niki Parmar, Jakob Uszkoreit, Llion Jones,
  Aidan~N Gomez, {\L}ukasz Kaiser, and Illia Polosukhin.
\newblock Attention is all you need.
\newblock {\em Advances in neural information processing systems}, 30, 2017.

\bibitem{CIDEr}
Ramakrishna Vedantam, C Lawrence~Zitnick, and Devi Parikh.
\newblock Cider: Consensus-based image description evaluation.
\newblock {\em Proceedings of the IEEE conference on computer vision and
  pattern recognition}, pages 4566--4575, 2015.

\bibitem{NOC}
Subhashini Venugopalan, Lisa Anne~Hendricks, Marcus Rohrbach, Raymond Mooney,
  Trevor Darrell, and Kate Saenko.
\newblock Captioning images with diverse objects.
\newblock {\em Proceedings of the IEEE conference on computer vision and
  pattern recognition}, pages 5753--5761, 2017.

\bibitem{SimVLM}
Zirui Wang, Jiahui Yu, Adams~Wei Yu, Zihang Dai, Yulia Tsvetkov, and Yuan Cao.
\newblock Simvlm: Simple visual language model pretraining with weak
  supervision.
\newblock {\em arXiv preprint arXiv:2108.10904}, 2021.

\bibitem{hugface}
Thomas Wolf, Lysandre Debut, Victor Sanh, Julien Chaumond, Clement Delangue,
  Anthony Moi, Pierric Cistac, Tim Rault, R{\'e}mi Louf, Morgan Funtowicz,
  et~al.
\newblock Transformers: State-of-the-art natural language processing.
\newblock {\em Proceedings of the 2020 conference on empirical methods in
  natural language processing: system demonstrations}, pages 38--45, 2020.

\bibitem{GroupViT}
Jiarui Xu, Shalini De~Mello, Sifei Liu, Wonmin Byeon, Thomas Breuel, Jan Kautz,
  and Xiaolong Wang.
\newblock Groupvit: Semantic segmentation emerges from text supervision.
\newblock {\em Proceedings of the IEEE/CVF Conference on Computer Vision and
  Pattern Recognition}, pages 18134--18144, 2022.

\bibitem{ShowAttendTell}
Kelvin Xu, Jimmy Ba, Ryan Kiros, Kyunghyun Cho, Aaron Courville, Ruslan
  Salakhudinov, Rich Zemel, and Yoshua Bengio.
\newblock Show, attend and tell: Neural image caption generation with visual
  attention.
\newblock {\em International conference on machine learning}, pages 2048--2057,
  2015.

\bibitem{ICSemanticAttention}
Quanzeng You, Hailin Jin, Zhaowen Wang, Chen Fang, and Jiebo Luo.
\newblock Image captioning with semantic attention.
\newblock {\em Proceedings of the IEEE conference on computer vision and
  pattern recognition}, pages 4651--4659, 2016.

\bibitem{Flickr30k}
Peter Young, Alice Lai, Micah Hodosh, and Julia Hockenmaier.
\newblock From image descriptions to visual denotations: New similarity metrics
  for semantic inference over event descriptions.
\newblock {\em Transactions of the Association for Computational Linguistics},
  2:67--78, 2014.

\bibitem{SMs}
Andy Zeng, Adrian Wong, Stefan Welker, Krzysztof Choromanski, Federico Tombari,
  Aveek Purohit, Michael Ryoo, Vikas Sindhwani, Johnny Lee, Vincent Vanhoucke,
  et~al.
\newblock Socratic models: Composing zero-shot multimodal reasoning with
  language.
\newblock {\em arXiv preprint arXiv:2204.00598}, 2022.

\bibitem{VinVL}
Pengchuan Zhang, Xiujun Li, Xiaowei Hu, Jianwei Yang, Lei Zhang, Lijuan Wang,
  Yejin Choi, and Jianfeng Gao.
\newblock Vinvl: Revisiting visual representations in vision-language models.
\newblock {\em Proceedings of the IEEE/CVF Conference on Computer Vision and
  Pattern Recognition}, pages 5579--5588, 2021.

\bibitem{OPT}
Susan Zhang, Stephen Roller, Naman Goyal, Mikel Artetxe, Moya Chen, Shuohui
  Chen, Christopher Dewan, Mona Diab, Xian Li, Xi~Victoria Lin, et~al.
\newblock Opt: Open pre-trained transformer language models.
\newblock {\em arXiv preprint arXiv:2205.01068}, 2022.

\bibitem{MemCap}
Wentian Zhao, Xinxiao Wu, and Xiaoxun Zhang.
\newblock Memcap: Memorizing style knowledge for image captioning.
\newblock {\em Proceedings of the AAAI Conference on Artificial Intelligence},
  34(07):12984--12992, 2020.

\end{thebibliography}
}

\clearpage

\section*{Supplementary Materials}

%-------------------------------------------------------------------------
\subsection*{A. Implementation Details}

After obtaining entities $\{{e_1}, {e_2}, {e_3}, ... \}$, we can construct the entity-aware hard prompt. To this end, we randomly drop certain entities (\textit{e.g.}, ${e_2}$) and insert the remaining ones into a prompt template, resulting in a sentence like ``There are ${e_1}, {e_3}, ...$ in the image.''
Subsequently,  we employ the tokenizer and word embeddings from GPT-2 to convert this sentence into dense vectors $\textbf{h} \in \mathbb{R}^{n \times d}$. Here, $n$ represents the length of vector $\textbf{h}$, and $d = 768$ indicates the dimension of GPT-2's latent space. The soft prompt, generated by the Transformer-based projector, is denoted as $\textbf{s} \in \mathbb{R}^{m \times d}$, where $m$ corresponds to the length of the soft prompt $\textbf{s}$.
Consequently, the prompt fed into GPT-2 can be represented as $\textbf{p} = \{\textbf{s};\textbf{h}\}, \textbf{p}\in \mathbb{R}^{(m + n) \times d}$, where $\{;\}$ denotes concatenation.

The auto-regressive objective is employed to train parameters $\theta$ of the decoder. It is defined as follows:
\begin{equation}
    \mathcal{L}_{obj} = - \frac{1}{|\textbf{w}|} \sum_{i = 1}^{|\textbf{w}|} \log p(w_i | \textbf{s}; \textbf{h}; {\textbf{w}}_{\le i}: \theta)
\end{equation}

We train ViECap on various source domains with the hyperparameters shown in Tab.~\ref{tab:traininghyperparameter}. During inference across different target domains, we retrieve visual entities using the frozen CLIP, which can be formulated as:
\begin{equation}
    p_{i} = \frac{\exp({\rm sim}(I, T_{i}) / \tau)}{ \sum_{j=1}^{N} \exp({\rm sim}(I, T_{j}) / \tau)}
\end{equation}
where ${\rm sim}(I, T_{i})$ denotes the cosine similarity between image $I$ and class name $T_{i}$, $\tau$ and $N$ refer to the temperature and the size of vocabulary, respectively. We choose the top $k$ class names with $p_{i}$ greater than threshold $p_{\rm thres}$ as retrieved entities. For all evaluations on cross-domain captioning, we leverage the same values of $k$, $p_{\rm thres}$, and $\tau$ (\textit{i.e.,} 3, 0.2, and 0.01, respectively). For evaluations on in-domain captioning, we set $k$, $p_{\rm thres}$, and $\tau$ to 3, 0.4, and 0.01 for COCO; 3, 0.3, and 0.01 for Flickr30k; 2, 0.1, and 0.007 for Flickrstyle10K.

\begin{table}[h]
\begin{center}
\small
\setlength{\tabcolsep}{1.5mm}{
\begin{tabular}{l|c|c|c}
\toprule
% \multirow{2}{*}{Method} & \multicolumn{4}{c|}{\textbf{Romantic}} & \multicolumn{4}{c}{\textbf{Humorous}}\\
\textbf{Hyperparameters} & \textbf{COCO} & \textbf{Flickr30k} & \textbf{FlickrStyle10K} \\
% ~ & B@1 & B@3 & M & C & B@1 & B@3 & M & C \\
\midrule
Epochs & 15 & 30 & 25\\
Batch size & 80 & 80 & 128 \\
Learning rate & $2e^{-5}$ & $2e^{-5}$ & $3e^{-4}$ \\
Masking rate & 0.4 & 0.4 & 0.4 \\
\bottomrule
\end{tabular}}
\end{center}
\caption{Training hyperparameter.}
\label{tab:traininghyperparameter}
\end{table}

%-------------------------------------------------------------------------
\subsection*{B. Unsupervised Metric}

\begin{table}
\vspace{-1em}
\small
\begin{center}
\setlength{\tabcolsep}{1.0 mm}{
\begin{tabular}{c|cccc|c|c}
\toprule
\multirow{2}*{Methods} & \multicolumn{4}{c|}{\textbf{COCO} $\Rightarrow$ \textbf{NoCaps val}} & \textbf{COCO} $\Rightarrow$ & \textbf{Flickr30k} \\
       & In & Near & Out & Overall & \textbf{Flickr30k} & $\Rightarrow$ \textbf{COCO} \\
\midrule
MAGIC  & 0.665 & 0.664 & 0.658 & 0.662 & 0.686 & 0.661  \\
CapDec & 0.711 & 0.701 & 0.671 & 0.692 & 0.737 & 0.694 \\
\rowcolor{Gray}
ViECap & \textbf{0.738} & \textbf{0.751} & \textbf{0.764} & \textbf{0.754} & \textbf{0.761} & \textbf{0.744} \\
\bottomrule
\end{tabular}}
\end{center}
\caption{Quantitative results in the cross-domain captioning using the unsupervised metric CLIP-S.}
\label{table:clips}
\end{table}

Furthermore, we report the captioning performance using the unsupervised metric, \textit{i.e.}, CLIP score (CLIP-S), to further validate the effectiveness of ViECap. We compare with other text-only methods (\textit{i.e.}, MAGIC, CapDec) in cross-domain captioning to assess the transferability of our model. As presented in Tab.~\ref{table:clips}, ViECap outperforms all other methods in cross-domain captioning by a large margin, indicating its robustness in handling domain shifts within diverse images.

%-------------------------------------------------------------------------
\subsection*{C. Hard Prompt Variants}

\begin{table*}
\small
\begin{center}
\setlength{\tabcolsep}{1.5mm}{
\begin{tabular}{l|c|cccc}
\toprule
\multirow{2}{*}{Hard prompt variants} & \multicolumn{1}{c|}{\textbf{COCO}}&\multicolumn{4}{c}{\textbf{NoCaps val}} \\
~ & Test & In & Near & Out & Overall \\
\midrule
\textbf{Default}: ``There are ... in the image." & 92.9 & 61.1 & 64.3 & 65.0 & 66.2 \\
\textbf{Variant 1}: ``There are ... in the scene. The image shows" & 92.6 & 59.2 & 63.5 & 64.2 & 65.2 \\
\textbf{Variant 2}: ``A photo of ..., a caption to describe this image is'' & 92.3 & 60.3 & 63.4 & 64.6 & 65.3 \\
\makecell[l]{\textbf{Variant 3}: ``To describe this image, let us think step by step. In this image, we can see ..., \\ so a sentence to describe this picture is''}
 & 91.5 & 59.2 & 62.7 & 64.9 & 64.9 \\
\bottomrule
\end{tabular}}
\end{center}
\caption{Results on variants of different hard prompt templates. ``..." denotes the parts to be filled by visual entities. }
\label{tab:hardpromptvariants}
\end{table*}

We explore the influence of different prompt templates on ViECap's captioning performance. As shown in Tab.~\ref{tab:hardpromptvariants}, ViECap shows minor sensitivity to changes in prompt templates, even when using a step-by-step hard prompt variant (variant 3). We speculate that the model is more effective for the altered parts in the template (\textit{i.e.,} visual entities ) due to fine-tuning GPT-2.

%-------------------------------------------------------------------------
\subsection*{D. Soft Prompt Length}

\begin{table}
\small
\begin{center}
\setlength{\tabcolsep}{1.5mm}{
\begin{tabular}{l|c|cccc}
\toprule
\multirow{2}{*}{Soft prompt length} & \multicolumn{1}{c|}{\textbf{COCO}}&\multicolumn{4}{c}{\textbf{NoCaps val}} \\
~ & Test & In & Near & Out & Overall \\
\midrule
Length: 10 & 92.9 & 61.1 & 64.3 & 65.0 & 66.2 \\
Length: 20 & 92.3 & 60.3 & 63.8 & 64.5 & 65.6 \\
Length: 30 & 92.3 & 60.8 & 63.9 & 65.3 & 66.0 \\
Length: 40 & 92.3 & 60.2 & 64.1 & 65.0 & 65.9 \\
\bottomrule
\end{tabular}}
\end{center}
\caption{Results on different lengths of soft prompts.}
\label{tab:softpromptlength}
\end{table}

We investigate the impact of different lengths of soft prompts on the captioning performance of ViECap. As shown in Tab.~\ref{tab:softpromptlength}, we arrive at the same conclusion as the experiment on hard prompt variants, \textit{i.e.}, increasing the length of soft prompts does not significantly improve the performance of ViECap while fine-tuning GPT-2.

%-------------------------------------------------------------------------
\subsection*{E. Time Cost}

\begin{table}
\begin{center}
\small
\setlength{\tabcolsep}{1.5mm}{
\begin{tabular}{l|c|c}
\toprule
\textbf{Models} & \textbf{Encoding + Retrieval} (ms) & \textbf{Decoding} (ms) \\
\midrule
ViECap & 20.39 + 0.57 & 127.99 \\
Faster R-CNN & 86.76 & - \\
\bottomrule
\end{tabular}}
\end{center}
\caption{The average time cost of captioning 100 COCO images using ViECap and Faster R-CNN during inference. Encoding denotes the time cost of encoding a single image to features. Retrieval refers to the average speed of detecting entities from features. Decoding refers to the average time cost of generating a sentence by the decoder.}
\label{tab:inferencespeed}
\end{table}

Tab.~\ref{tab:inferencespeed} compares the time cost of CLIP-based retrieval and detector-based retrieval (\textit{i.e.}, Faster R-CNN). We calculate the average time cost of processing 100 images from the COCO testing set on a single NVIDIA TITAN V GPU. For detector-based retrieval, we use Faster R-CNN with the backbone of ResNet-101\footnote{We utilize the model and pre-trained weights from~\url{https://github.com/open-mmlab/mmdetection}}. The results indicate that our model is four times faster than Faster R-CNN, from processing a single image to obtaining the detected entities. Note that the integrating of additional entity-aware hard prompts only incurs a minor time increase of 0.57 $ms$ compared to CapDec while significantly outperforming CapDec by a large margin across various benchmarks.

%-------------------------------------------------------------------------
\subsection*{F. Vocabularies}

The quality of the vocabulary impacts the retrieval performance of CLIP and the transferability of ViECap. For results reported in this paper, we leverage the COCO vocabulary for the COCO testing set and the VGOI vocabulary for all other datasets.
Visual Genome contains various class name annotations, but they suffer from noise and incorrect annotations. We select class names consisting of a single word to construct Visual Genome vocabulary (17069). Zhang \textit{et al.}~\cite{VinVL} clean Visual Genome to build a clean corpus (\textit{i.e.}, VGOI vocabulary), which comprises 1848 class names. We also construct the COCO (80) vocabulary and the Open Image (601) vocabulary using class names from the corresponding class annotations.

The NoCaps dataset contains three domains: 1) \textit{in-domain} only contains COCO classes, 2) \textit{near-domain} contains both COCO and Open Image classes, and 3) \textit{out-of-domain} only contains Open Image classes.
Tab.~\ref{tab:vocabularies} shows the results of NoCaps on different vocabularies. A specific domain of the captioning dataset benefits from a specific vocabulary (\textit{e.g.}, COCO vocabulary achieves the best performance in the \textit{in-domain} of NoCaps, and Open Image vocabulary achieves the best performance in the \textit{out-of-domain} of NoCaps).
However, when aiming for transferability to a novel domain where a specific vocabulary is not attainable, a large, diverse, and clean vocabulary describing various classes becomes crucial. As shown in Tab.~\ref{tab:vocabularies}, the VGOI vocabulary achieves a great trade-off between the \textit{in-domain} and \textit{out-of-domain} captioning performance. Notably, a large but noisy vocabulary, as seen in the Visual Genome vocabulary in Tab.~\ref{tab:vocabularies}, does not significantly improve ViECap's performance. 

\begin{table}[h]
\small
\begin{center}
\setlength{\tabcolsep}{1.2mm}{
\begin{tabular}{l|c|cccc}
\toprule
\multirow{2}{*}{Vocabulary} & \multirow{2}{*}{Size} & \multicolumn{4}{c}{\textbf{NoCaps val}} \\
~ & ~ & In & Near & Out & Overall \\
\midrule
COCO vocabulary & 80 & 63.6 & 51.0 & 22.7 & 44.9 \\
Open Image vocabulary & 601 & 59.5 & 66.8 & 69.4 & 69.2 \\
VGOI vocabulary & 1848 & 61.1 & 64.3 & 65.0 & 66.2 \\
Visual Genome vocabulary & 17069 & 56.8 & 50.5 & 41.9 & 50.0 \\
\bottomrule
\end{tabular}}
\end{center}
\caption{Results on NoCaps using different vocabularies.}
\label{tab:vocabularies}
\end{table}

%-------------------------------------------------------------------------
\subsection*{G. Datasets}

COCO and Flickr30k are commonly used benchmarks for evaluating image captioning models. We divide these datasets into three parts (\textit{i.e.}, training, validation, and testing set) following the Karpathy \textit{et al.} split~\cite{alignmentcaption}. This results in 113,000, 5,000, and 5,000 samples for COCO and 10,300, 1,000, and 1,000 samples for Flickr30k, respectively.

NoCaps is divided into three domains, evaluating the capability of models to describe novel objects in images - \textit{in-domain} consists solely of COCO classes, \textit{near-domain} includes both COCO and novel classes, and \textit{out-of-domain} comprises only novel classes. As suggested by OSCAR~\cite{Oscar}, we assess the models using only the validation set.

Additionally, FlickrStyle10K assesses the task of generating captions with new styles, \textit{i.e.}, ``romantic'' and ``humorous''. Since only 7,000 training samples are publicly available, following the approach used in MemCap~\cite{MemCap}, we randomly sample 6,000 captions as our training set, while the remaining image-text pairs constitute our testing set. 

%-------------------------------------------------------------------------
\subsection*{H. Visualizations}

\begin{figure*}
\begin{center}
   \includegraphics[width=0.9\linewidth]{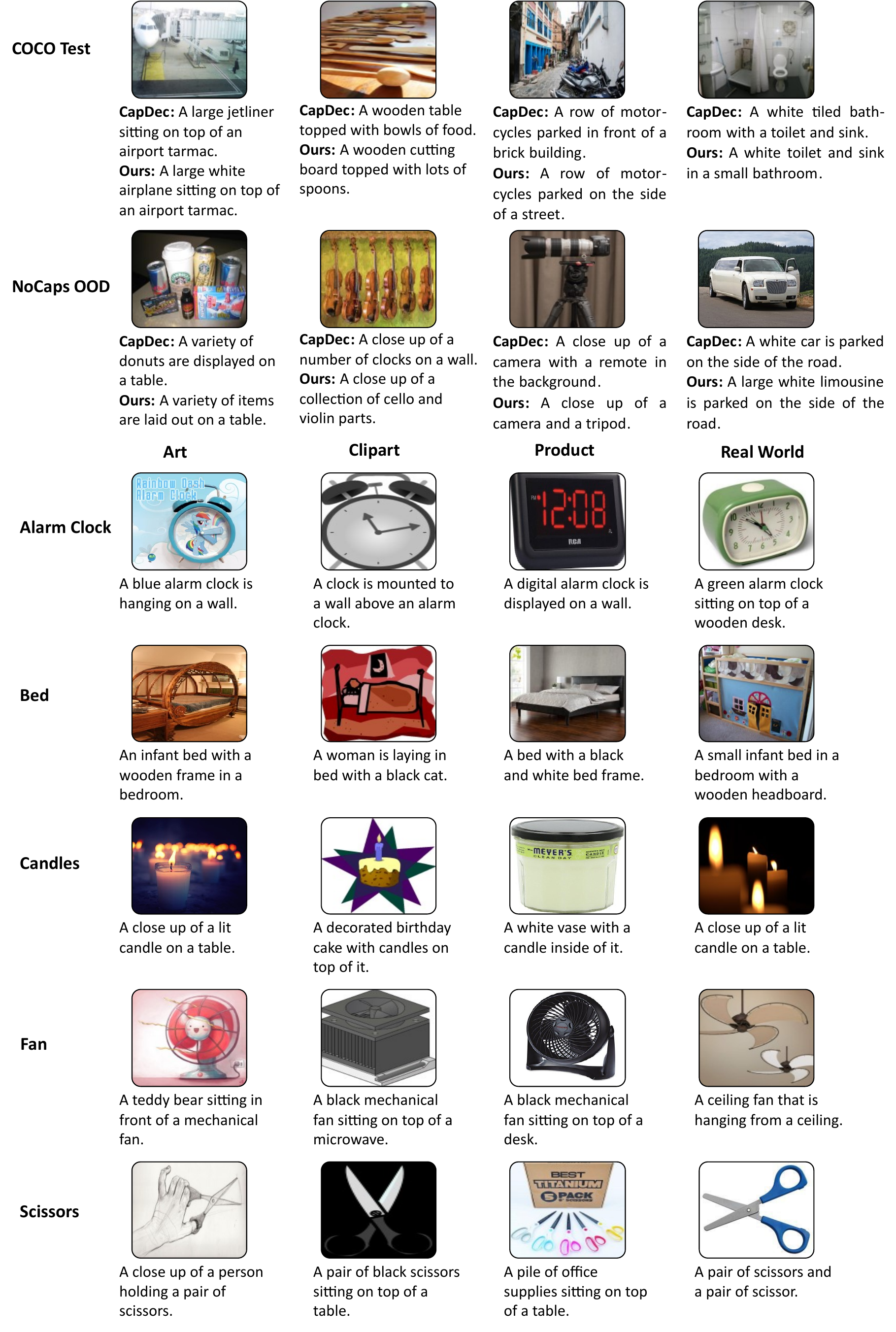}
\end{center}
   \caption{More visualization results of ViECap on in-domain captioning (row 1), cross-domain captioning (row 2), and image-domain-adaptation captioning (row 3-7).}
\label{fig:morevisualization}
\end{figure*}

Additional visualization results are presented in Fig.~\ref{fig:morevisualization}, showcasing the remarkable transferability of ViECap. Our model excels not only in describing novel objects but also in generating captions for images with various styles.

Here, we leverage weights trained on the COCO training set for captioning. The first row displays the captioning results on the COCO testing set, demonstrating the successful description of in-domain objects by both CapDec and ViECap.
The second row presents results on the \textit{out-of-domain} of NoCaps, showcasing ViECap's ability to generate high-quality texts related to unseen objects.

Rows 3 to 7 illustrate captioning results for Office-Home~\cite{OfficeHome}, a benchmark dataset for image domain adaptation, which comprises four different styles of image domains: 1) Art, artistic images in the form of sketches, paintings, ornamentation, \textit{etc}., 2) Clipart, collection of clip art images, 3) Product, images of objects without a background, and 4) Real-World, images of objects captured with a regular camera.
We evaluate the captioning performance of ViECap across these diverse image styles, using the first image from different categories in Office-Home (\textit{i.e.}, we do not choose a specific image but simply use the first image of each class in the dataset). Despite a few incorrect captions, ViECap is capable of describing different styles of images with reasonable descriptions in most cases, highlighting that our captioning model can effectively transfer to various styles of images and generate appropriate captions related to images.

\end{document}